# MACHINE LEARNING AND THEORY-LADENNESS: A PHENOMENOLOGICAL ACCOUNT

Alberto Termine, SUPSI, IDSIA, Lugano

Emanuele Ratti[1], University of Bristol

Alessandro Facchini, SUPSI, IDSIA, Lugano & MINDS, Kozminski University, Warsaw

**Abstract.** We provide an analysis of theory-ladenness in machine learning (ML) in science, where 'theory' (that we call 'domain-theory') refers to the domain knowledge of the scientific discipline where ML is used. By constructing an account of ML models based on a comparison with phenomenological models, we show (against recent trends in philosophy of science) that ML model-building is mostly *indifferent* to domain-theory, even if the model remains theory-laden in a weak sense, which we call *theory-infection*. These claims, we argue, have far-reaching consequences for the *transferability* of ML across scientific disciplines, and shift the priorities of the debate on theory-ladenness in ML from *descriptive* to *normative*.

## 1. INTRODUCTION

The development of data-intensive methods in the sciences (from Big Data, data science, to AI) has been often associated with the idea that these can function without inputs from scientific expertise, or without appealing to domain knowledge of the scientific fields in which they are used. This idea was central in the debate on the so-called 'two cultures of statistical modeling' (Breiman 2001). Famously, Breiman distinguishes two ways of doing statistics, one more "traditional" (called 'data modeling culture') based on the adoption of data-models that statisticians formulate with the help of domain-related theoretical considerations, and one relying on far less assumptions and based on learning predictive algorithmic models directly from data with few constraints (called 'algorithmic modeling'). The way Breiman describes the 'algorithmic modeling culture' overlaps significantly with the strategies associated with concepts like 'data-intensive scientific discovery' (Hey et al 2009), 'data-driven' science, Big Data, machine learning, and now, more in general, AI[2]. In all these cases, there is an emphasis on leveraging data by relying on a few 'neutral' constraints given by 'general-purpose' tools that can be easily adapted to a variety of application domains and tasks, and increasingly less on domain-specific assumptions. The prototypical example of this tendency are arguably multi-modal foundation models, which are increasingly common in contemporary AI research. These are large deep neural networks pre-trained on huge corpora of data (including images, texts, audio etc.), and conceived to be easily adaptable to a variety of tasks via a simple fine-tuning or even with just a proper prompt through an API.

[1] Corresponding author, mnl.ratti@gmail.com

[2] Such independence has been even popularized as a new scientific paradigm which came to be known, infamously, as the 'End of Theory' (Anderson 2008)

Gauging whether data-intensive methods are independent from, or less reliant on, theoretical considerations coming from the scientific domain of implementation means asking a question about the theory-ladenness of data-intensive methods. By 'theory-ladenness', we mean here the idea that, in data-intensive science, engaging in essential scientific *activities* requires either the implicit assumption of, or the explicit appeal to, scientific theories. We understand the term 'theory' in a broad sense (specified in more detail below) to include *domain knowledge and expertise* of the given scientific field in which data-intensive methods are used. While in data-intensive science - and especially in machine learning (ML) - there might be ideas coming from statistical learning theory that can reasonably count as 'theory', here theory-ladenness is *restricted to the 'domain-theory' belonging to the scientific context of implementation of the model*.

With few exceptions (e.g., Napoletani et al. 2020), philosophers of science have been systematically arguing in favour of theory-ladenness (Callebaut 2012; Kitchin 2014; Leonelli 2016; Boon 2020; Knusel and Baumberger 2020; Hansen and Quinon 2023), by pointing to its *inevitability*, and showing the subtle ways in which theoretical considerations (broadly conceived) inform the construction and use of these data-intensive methods.

The aim of this article is to argue *against the inevitability* of traditional forms (as analysed in the philosophical literature) of theory-ladenness when applied to the specific case of ML[3]. In particular, we argue that *practices* related to the construction of ML models (MLM) are *theory-indifferent*, which is a term we create (and explain in great detail) to indicate that no explicit reference to the domain-theory (i.e., the theory of the specific domain in which a MLM is implemented) is necessarily required in the modelling practices. This is true despite the fact that MLMs themselves are theory-laden - and they are so in a way that is significantly different (and arguably weaker) than any other form of theory-ladenness discussed in the literature (we call this *theory-infection*). Asking the question of theory-ladenness (and arguing against the inevitability of theory-ladenness) has a number of implications, as we will show. If ML-based data-intensive methods are indeed theory-indifferent, then data scientists/AI practitioners do not need much information coming from scientific expertise, and their tools and methods can flexibly travel across different scientific domains. If data-intensive science based on, say, cutting-edge AI methods, need not be theory-laden, then the training of future scientists should be especially focused on domain-agnostic and engineering aspects of data-intensive methods, rather than highly discipline-specific curricula (e.g., computational biology).

[3] The reason for focusing on ML is that it is the *fil rouge* in the different labels that have been given to data-intensive strategies (such as algorithmic modeling, data-driven science, Big Data, data science, AI, etc). For instance, Breiman cites support vector machines, decision trees, and neural nets. In Hey et al (2009), machine learning techniques are mentioned in several examples of the 'data-intensive' paradigm, such as in neurobiology, healthcare, etc. What have now been rebranded as 'AI techniques' include known machine learning techniques such as deep neural networks, support vector machines, and others. How exactly ML emerged as being central in all these different strategies and contexts is a topic that it is yet to be addressed properly in the history of science and technology

The structure of the article is as follows. After a brief introduction on the *status quaestionis* of theory-ladenness of ML and a precise characterization of terms like 'theory' (Section 2), we show that the investigation of this topic requires an analysis of the interactions between ML and scientific domain knowledge based on a precise account of what MLMs are. We construct an account of MLMs by comparing them to phenomenological models (PMs), and we show that this comparison can illuminate the role that theory plays in ML-based science (Section 3). Because of their similarities with PMs, MLMs can have a high-degree of what we call *theory-indifference* in the way they are *constructed* (Section 4), that is, neither a reference to domain-theory, nor scientific-domain expertise is necessary in the various steps that undertake the *construction* of MLMs. This remains true even if MLMs, *per se*, are *theory-infected*, which is a peculiar way of being theory-laden that is different from the various existing forms of theory-ladeness of scientific models discussed in the philosophical literature. This, we argue, has two specific implications (Section 5). First, our analysis sets ML modelling practices apart from more traditional modelling methods, especially for what concerns the nature of the 'transferability' of ML methods across contexts. Second, the debate on the theory-ladenness of ML can shift from descriptive (i.e., how is ML theory-laden?) to normative goals (i.e. should ML be theory-laden?).

## 2 MACHINE LEARNING AND THEORY-LADENNESS

There is a notable trend in philosophy of science according to which fundamental aspects of scientific practice are theory-laden. By 'theory-ladenness', here we mean the idea that it is not possible to engage in specific scientific activities without appealing to a number of theoretical considerations, both implicit and explicit (Boon 2020; Longino 2020). Whether it is meant that everything is necessarily theory-laden, or just possibly theory-laden is unclear, but works on experimentation and modelling have often given the impression that theory is always present, especially in subtle ways. We call this the *blanket view* of theory-ladenness. In order to characterize the blanket view more precisely, we want to be clear on what we mean by ‘theory’, and what we take to be representative examples of ‘theory-ladenness’.

We understand ‘theory’ along the lines of the characterization provided by Douglas and Magnus (2013) and further developed by Ratti (2020). Douglas and Magnus (2013) distinguish four levels across which scientists make inferences. Data is the first level, where ‘data’ is understood broadly to encompass traditional notions (e.g., Bogen and Woodward 1988), as well as more recent accounts (e.g., Leonelli 2016). Phenomena is the second level, and here as well this notion is understood along the lines of Bogen and Woodward’s characterization (1988). The third and the fourth level - the ones we are especially interested in here - include ‘theory’, and ‘framework’ respectively.

Douglas and Magnus (2013) understand ‘theory’ as a set of models and laws that explain and predict a broad class of phenomena pertaining to a specific domain, while the ‘framework’ refer to a set of assumptions, auxiliary hypotheses, and theoretical commitments that characterize the specific domain to which models and laws apply. In the context of, e.g., mechanistic sciences like cell biology, neuroscience, or chemistry, ‘theory’ could be understood as the set of the mechanistic models that are used to explain natural phenomena within a specific scientific domain. ‘Framework’ can be understood in various (often equivalent) ways. One can conceptualize ‘framework’ as the ‘theoretical’ components of the toolbox of science, which offer “the tools for constructing representations” provided by models (Suarez and Cartwright 2008, p 65). An alternative formulation of ‘framework’ relies on the notion of ‘store of the field’ (Darden 2006), understood as a set “of established and accepted components out of which mechanisms [i.e., mechanistic models] can be constructed” (p 51), as well as accepted modules, namely “organized composites of the established entities and activities” (p 51) that are relevant to construct models. For instance, in cell biology examples of components include DNA or RNA molecules, activities include phosphorylation or acetylation, and modules might be ribosomes. Another way of understanding ‘framework’ is by using Longino’s concept of ‘explanatory model’, which is a characterization of the items that are contained in scientific explanations, and the relationships between them (1990, p 134). This can include a number of different components, from auxiliary assumptions to highly specific terms and ways of using them within a given scientific context. In this article, the ‘theory’ of ‘theory-ladenness’ is broadly conceived to include the third and the fourth levels of Douglas and Magnus’ account (namely, ‘theory’ and ‘framework’). This means that ‘theory’ is not just the sum of all models and/or laws of a given scientific domain, but it also includes the theoretical commitments, auxiliary assumptions, and the vocabulary used to talk about the phenomena of that domain. *From now on, we use the term 'domain-theory' to refer to both Douglas and Magnus' notions of 'theory' and 'framework'* (unless otherwise specified).

Turning to theory-ladenness, and without claiming to exhaust a complex and multifaceted debate, we think that there are three dominant senses in the literature. First, there is a sense in which scientific activities like experiments can be designed to test a given theory. In such cases, theory-ladenness is *theory-testing*, and the connection between domain theories and scientific practice is obvious. But for this article, two other distinct senses are more relevant: theory-ladenness intended as *theory-directedness,* and as *theory-informed*. On the one hand, there is theory-directedness when the theoretical background directs inquirers to properties of scientific phenomena that could possibly play a causal role in investigations or, in alternative, when a local theory will give you reasons to expect that certain things will happen in a given experiment or, more generally, in a scientific activity (Franklin 2005). Similarly, Waters (2007) takes theory-directedness to have two forms: a 'weak' form where one theory directs scientific activities to look for certain facts and not others; and a 'strong' form where a theory generates expectations about what will be observed. On the other hand, theory-informed is when a theory is used, explicitly (though one might not be necessarily aware of it), to provide background and constraints on how science is practised (Elliott 2007). Radder (2003) uses other terms, but these can be categorized as either 'theory-directed' or 'theory-informed'. For instance, 'theory-guidance' is a case of theory-directedness, given the role played by theory in directing successive scientific activities. Similarly, close to theory-directedness is the idea that a scientific activity acquires its significance from its direct bearing on disputes which might be theory-laden themselves or simply be about competing theories. Finally, he also mentions that often a theoretical background provides constraints on how to read data. This can be interpreted as 'theory-informed'.

Now, the blanket view of theory-ladenness is the idea that in order to engage in scientific activities, a commitment to domain-theory is *necessary*. This commitment can be in a strong form like theory-testing or theory-directedness, or in subtler forms such as 'theory-informed'. But in all cases, once we look at how the science is done, *explicit* references to aspects which can be traced back to domain-theory (understood both as 'theory' or 'framework') will be found and, sometimes, these are seen as necessarily required to practice science. We see this at play in discussions on experiments and modelling activities.

In the context of experimentation, there has been an explosion of analyses supporting the blanket view. For instance, exploratory experiments have been seen as only loosely guided by domain-theories (Steinle 1997). This could be expressed especially by seeing this class of experiments as theory-informed (Waters 2007; Elliott 2007), or that theoretical interpretation is necessary for executing a given experiment (Radder 2003). In the terminology introduced above, the idea is that exploratory experiments necessarily make use of the 'framework' (or store of the field, or explanatory model, or 'theory as a toolbox'). This is unlike stronger theory-directed experiments (Waters 2007), where "a theory generates expectations about what will be observed" (p. 277). In this case, theory is assumed in the strict sense of 'theory' as the third level of Douglas and Magnus' hierarchy, since a 'model' or a 'hypothesis' as it appears in the set of models of the theory is scrutinized, and the outcomes of an experiment are evaluated on its basis. What this literature has tried to establish is that scientific activities like experimentation cannot possibly be independent from domain-theory (Radder 2003).

The strict connection between models and theories has been a topic of interest at least since the early formulations of the semantic view of theories. In addition to traditional ideas where models are derived directly (and solely) from theories, more recent views recognize that the construction of models at least requires the appeal to components coming from domain-theories. A classic strategy is to say that the choice of parameters, variables, model descriptions, and structures (Weisberg 2013), as well as metrics for what counts as a successful output, etc, are based on a pre-existing understanding of the phenomena to investigate, of the scientific goals, and based on scientific norms that seem to rely significantly on domain-theory. In this case, one can say that models are at least theory-informed. In the debate on `models as mediators' (Morgan and Morrison 1999), even if models maintain a partial independent status from both theory and data, models are still theory-mediated, since models "typically involve some of both [i.e. theory and data]" (p. 11), where 'involving theory' can be understood along the lines of theoretical commitments of the forms described above. There are also stronger cases: 'traditional statistical modelling' (what Breiman refers to as the "data-modelling culture") is, indeed, not only theory-informed by the framework, but theory-directed in a number of important ways (Shmueli 2010). In fact, Cox argues that choosing the right statistical model to start an investigation requires the translation of "a subject-matter question into a mathematical and statistical one, and clearly the faithfulness of the translation is crucial" (2006, p 3). Therefore, even in the case of models the question is not whether they are theory-laden, but rather *how,* and *which role* domain-theory plays[4].

With few exceptions (Napoletani et al. 2020), the 'blanket view' trend seems to be leading philosophical discussions on ML. A significant number of works have been making

[4] The only exception, as we will see, comes from the discussion on phenomenological models

the claim that ML methods, and/or MLMs, are at least (domain) theory-informed. Sometimes the claim of theory-ladenness is particularly weak, as in the case of Pietsch (2015), who shows that, even if ML methods can be *internally* theory-free, they still are *externally* theory-laden (even though the consequences of this for the structure of ML systems is not explored in detail). Knusel and Baumgartner (2020) entertain the idea that 'data-driven' models can retain a level of independence from scientific domain knowledge, but in the end they seem to claim that domain-theory is needed to build better models, even though it is not clear how. However, in other cases works still seem to imply that ML is *inevitably* theory-laden. For instance, Callebaut (2012) argues that Big Data methods (and hence what are now known as ML methods) require significant appeals to scientific perspectives, which usually come from domain-theories. Kitchin (2014) argues that domain-theory is inevitable, even though he does not distinguish neatly between scientific domain knowledge and engineering practices. Boon (2020) has argued that empiricists fantasies of a science *theory-empty* in ML are just doomed to fail, because every "tiny step in these intricate research processes involve epistemic task (...) for which all kinds of practical and scientific knowledge is crucial" (p 61). Hansen and Quinon (2023) argue that "theoretical background is involved in data generation, problem formulation, and algorithm evaluation" (p. 16). Moreover, they also argue that even 'engineering' activities like the construction of model architectures necessarily require domain-theory. Andrews (2025) shifts back and forth between the idea that ML should be theory-laden and stronger claims that ML is necessarily theory-laden. Ratti (2020) stresses the impossibility of using MLMs in biology without resorting to an interpretation that is shaped by domain-theory. Finally, Gross (2024) shows a strict interplay between MLM construction and mechanistic approaches in biology, where the 'mechanistic approaches' are, indeed, laden with the so-called 'store of the field'. What is important about these examples is that theory-ladenness is assumed, and the conceptual work left to do for philosophers is to identify theory-laden facets of ML model components or practices, spanning across the third and fourth level of Douglas and Magnus' hierarchy.

In what follows, we show that this blanket view is problematic, and that this has important implications for the nature of ML as a modelling strategy, as well as for the nature of the debate on theory-ladenness in ML-assisted science.

## 3. MACHINE LEARNING MODELS: A PHENOMENOLOGICAL ACCOUNT

In order to grasp the relation between domain-theory and MLMs, it is important to characterize more precisely what MLMs are in the first place. In this section, we first clarify the meaning we attribute to the term 'model' in ML, and then provide an account of MLMs based on a comparison with so-called *phenomenological models* (PMs).

### 3.1 Machine Learning Models and Machine Learning Systems

In literature, MLM may be used quite ambiguously to denote a variety of different things. For example, the term MLM is often used as a synonym for ML algorithm, i.e. an algorithm capable of learning patterns from data (see e.g., Bishop 2006). These algorithms produce data-fitting curves, which are also commonly referred to as MLMs, as are the computer architectures that implement such algorithms. To solve this terminological ambiguity, we propose to make a fundamental and explicit distinction between the term MLM and the term *ML system*. The latter is used broadly to refer to any computational artifact capable of learning information from data and adapt its behaviour accordingly. ML systems share all a general architectural framework, which encompasses three main modules (Facchini & Termine 2022):

1. the training sample
2. the training engine
3. the learned model.

The *training sample* is the repository of observational/synthetic data that the system uses as the source of information to learn and adapt its behaviour. The individual *data-points* of this sample denote specific instances of the system's target-phenomenon of interest and are composed of *features*, i.e., mathematical representations encoding specific measurable magnitudes of the target-phenomenon of interest (for example, the *colour of a given pixel* in an image or the *age* of a given patient in a biomedical study). Selecting and constructing the proper features have a crucial impact on the predictive performances of a ML system, because predictions are generated by analysing the correlations between the single features in the training sample and occurrence, or the probability of occurrence, of the specific target-phenomenon. This process is usually referred to as *features engineering* and requires the analysis of different possibilities and a suitable combination of statistical techniques. The process starts with sampling relevant properties of a target-phenomenon from pre-processed data, which are thus mapped into measurable variables called *raw-features*. The latter are further processed through the iterative application of several data-transformations, which eventually lead to *derived features* (also called *embeddings*) better suitable for prediction. More specifically, the process of selecting and constructing derived features can be 'hand-made' or performed automatically through appropriate *feature learning algorithms*, this being mostly the case for advanced contemporary ML systems, such as deep neural networks (Baldi 2021).

The *training engine* is the computational machine that allows a ML system to 'learn' from its training data. This machine implements a ML algorithm, i.e., a computational procedure that performs an iterative adjustment of the system's input-output behaviour with the goal of minimising the system's prediction error, and thereby improving its predictive performance at test time. Technically, the training engine performs this iterative adjustment by updating a vector of parameters *w* that governs the overall system's behavior. This updating is driven by a function *f*, called the *objective function*, related to the parameters vector *w* and that measures the system's predictive error over the training sample. The goal of the training engine is thus to minimise/maximise the function *f* by iteratively tuning the parameters in *w* via a suitable optimisation procedure. Both the specific nature of *f* and of the optimisation procedures used to maximize/minimize it varies depending on the *learning paradigm* adopted (e.g., supervised learning, unsupervised learning, reinforcement learning, self-supervised learning). For example, in supervised learning, *f* is usually a *loss function* that measures, for each instance *x* in the training sample, the distance between the correct prediction *y*, which is assumed to be known for each instance *x* in the training sample, and the prediction *m(x)* that the system *m* outputs for *x*. This distance has usually to be minimized. A common example of loss function is the *mean squared error* of the system, which measures the distance between *y* and *m(x)* as the (average) squared difference between *y* and *m(x)*.

Notice that, although ideally the goal of training would be to find the *global optimum* (minimum/maximum) of the objective function, this is typically impossible in practice due to the highly-complex non-convex nature of *f*, which makes the search for the global optimum computationally unfeasible. Hence, existing techniques focus on the search of *sufficiently good* local minima of *f*, that is, in very general terms, local minima that allow the system to achieve practically acceptable levels of predictive error. This search is performed via suitable heuristic strategies that balance accuracy with scalability and computational feasibility. There exist a great variety of such heuristic strategies. One of the most widely adopted, especially in *deep learning*, is the *stochastic gradient descent*, which is based on the computation of the gradient of the loss function with respect to the parameters *w*. The procedure exploits a basic concept of differential calculus, namely the equivalence between the partial derivative and the slope degree of the tangent line to a function at each of its points. As one approaches the point of minimum (see Fig. 1), the derivative will tend to decrease (i.e. the tangent line gradually decreases its slope) until it reaches a point of minimum.

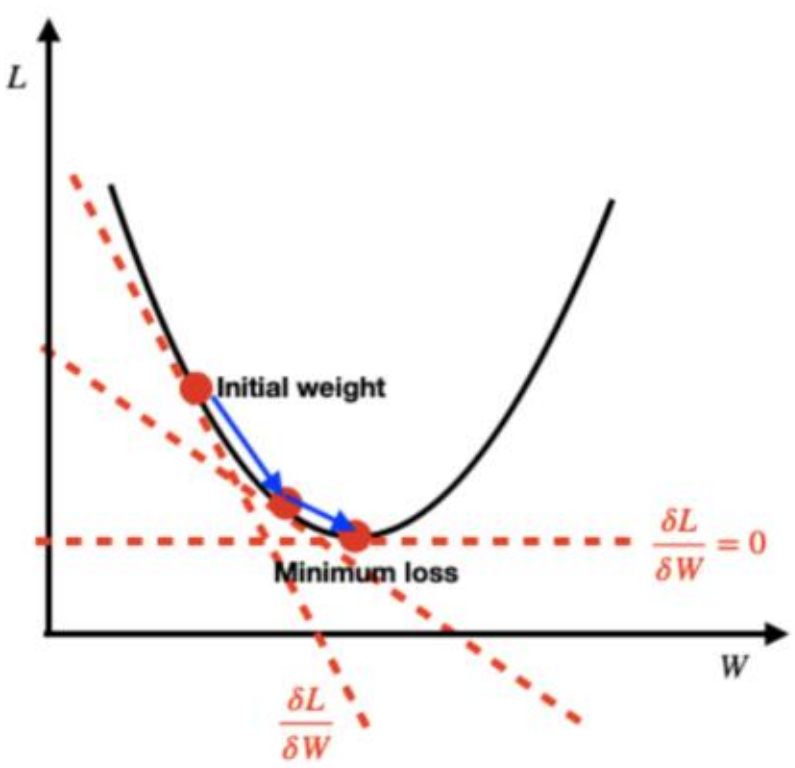

**Figure 1**: A simplified representation of the gradient descent procedure obtained by assuming the loss function to be defined over a single-element parameter vector. The curve represents the value of the loss for different values of the weights. Dotted lines represent the tangents of the loss for different values of the gradient of the loss with respect to the weight. Please note that, to facilitate understanding, we assume a convex parabolic loss defined over one single parameter. In practice, loss functions are typically highly-complex non-convex functions defined in highly-dimensional spaces, which may encompass millions or even billions of parameters. For more details on loss-functions, their nature and role, see Bishop 2006, Ch.4. We thank a reviewer for pressing on this point

The third component of an ML system we consider in our analysis is what we call the *learned model*. This is a mathematical representation of the statistical patterns that the system learns from data and uses to formulate predictions on the target-phenomenon of interest. At an abstract level, this representation is a *fitting curve* defined in a *n*-dimensional space, called *feature space*, whose axes codify the values of the features[5]. This curve is what scientists commonly denote with the term *MLM* in the context of scientific research. Understood in these terms, a MLM seems not substantially different from more 'traditional' examples of statistical models used in scientific practice, such as linear or logistic regression (Dobson & Barnett 2018)[6]: both kinds of models are essentially data-fitting curves. However, there are important differences between `traditional' statistical models (as in Breiman's 'data-modelling culture') and MLMs.

The first difference concerns *who* fits the data (i.e., the agent responsible for finding the data-fitting curve): in traditional statistical modelling practices, the fitting of data is mostly an *hand-made* task performed by a domain-expert with extensive statistical competences (such as a *bio-statisticians*, or *an expert in statistics for econometrics* etc.). In ML-based modelling, the operation of fitting the data is completely *automated* and the 'agent' responsible for it is the training engine. As we will clarify more in detail in the following sections, this has fundamental consequences for theory-ladenness of MLMs compared to that of traditional statistical models.

[5] Please note that this function can be practically implemented, at the level of the algorithm architecture design, in different formats. For example, a *linear classifier* can be equivalently implemented both by a decision-tree, a neural network, a support-vector machine etc. This is a consequence of the fact that computational models (including MLMs) can be represented at different *levels of abstraction* (on this point, see, Floridi 2008, Angius et al. 2021, Primiero 2019, Facchini & Termine 2022). But in the specific context of scientific research, what is relevant is the most abstract level of abstraction, where all MLMs can be represented as fitting curves in the features space.

[6] Note that linear regression can be considered either as a traditional statistical model or as an MLM. The difference between a traditional linear regression and a linear regression understood as an MLM lies in the way the parameters and hyperparameters of the model are specified, as we explain below.

Another relevant point of difference concerns the high dimensionality of MLMs compared to traditional statistical models (this is what Breiman 2001 refers to as the 'Occam problem'). The term 'dimensionality' has a very specific meaning in this context, i.e., it refers to the *number of dimensions* of the model's feature space. In 'traditional' statistical models, the number of features (i.e., variables) considered is limited and this allows these models to be easily represented graphically as lines or planes in low-dimensional spaces. For contemporary MLMs, and particularly in deep learning, the number of features considered is typically very large, making it impossible to represent the learned models in a suitable and humanly understandable manner (see Fig. 2).

A third difference is a consequence of the second. The difficulty of representing MLMs in suitable graphical (e.g., curves in a plane) or analytic (e.g., linear equations) forms because of high dimensionality, severely limits their transparency and interpretability (Selbst and Barocas 2018)[7]. To clarify this issue, consider a simple example (Fig. 2). Take a traditional linear regression model that analyses the correlation between *age* and *cancer risk*[8]. This model can be easily represented graphically as a curve in a two-dimensional plane (Fig. 2a), or analytically as a linear equation $Y = rX + b$ (where $r$ measures the "relevance" of $X$ for $Y$). Both these two formats of representation make it easy for scientists to grasp the statistical information the model embeds: one has just to observe the slope of the regression line in Fig. 2 to realize that the model identifies a *positive correlation* between the feature *age* and the target-phenomenon *cancer risk* (the greater the age, the greater the cancer risk). Similarly, it is sufficient to observe the *weight* (parameter $r$) of the variable $X$ (representing *age*) to understand the degree of statistical correlation existing between this feature and the target-phenomenon of interest. Now consider the graphical representation of an MLM provided in Fig. 2b. In this case, it is challenging (or, we can say, impossible) to capture any statistical pattern between features and target-phenomena by looking at this type of graphical representation, since it is far too complex. As we shall see in the course of the paper, these issues have profound epistemological implications for the integration of MLMs in scientific practices and mark a substantial gap between the latter and other types of statistical models commonly used in scientific research.

[7] This issue has been widely discuss in the philosophical literature, especially within the debate on the *opacity* of MLMs (Burrel 2016, Duran and Formanek 2018, Paez 2019, Creel 2020, Sullivan 2022, Zednik 2021, Zednik and Boelsen 2022, Boge 2022, Facchini and Termine 2022).

[8] In some cases, linear regression models can be also considered a simple instance of MLMs, notably when their parameters are learned automatically via optimisation procedures. However, we are focusing in this example on the case of a more 'traditional' linear regression model, whose parameters are manually fitted.

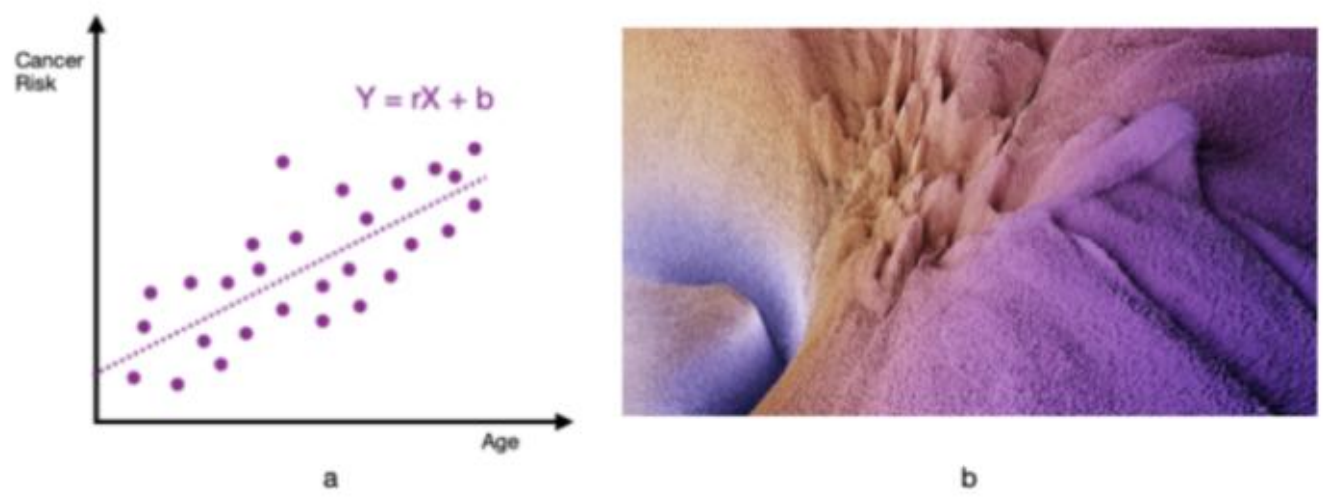

**Figure 2**: A graphical representation of a *linear regression model* (a) compared with a low-dimensionalised graphical representation of a MLM learned by a deep neural network (b). Image (b) borrowed from https://losslandscape.com/gallery/.

### 3.2 Machine learning models and phenomenological models

In the Introduction we mention that seeing MLMs through the lens of phenomenological models (PM), can shed light on the intricate theory-ladenness dynamics of ML practices. Now, in which sense PMs can provide a useful lens to understand what a MLM - i.e., the *learned model* as described in Section 3.1 - is really about? Let us start by clarifying the nature of PMs first.

In philosophy of science, PMs have been seen as either models of phenomena, or models that are not-theory-driven, or simply as models derived from measurements (Cartwright and Suarez 2008). Underneath these disagreements, philosophers of science tend to agree on four basic characteristics of PMs.

The first characteristic (C1) is to be found in an article by McMullin (1968), where he distinguishes between theoretical laws or explanations, and PMs. A theory, in his account, is not just a description of the evidence, but it goes beyond the evidence by entertaining the existence of “a postulated physical structure that could provide a causal account of the data to be explained” (1968, p. 388). The theoretical model is the (representation of the) postulated structure. A PM is different: it is “an arbitrarily-chosen mathematically-expressed correlation of physical parameters from which the empirical laws of some domain can be derived” (p. 391). As such, PMs account for evidence “in convenient [mathematical] form” (p. 391), but they do not postulate any physical structure, like theoretical models[9]. As an example, he considers a data set of cosmic ray showers. In order to bring the data into a single array, one can just hypothesize that they follow a general distribution function of, say, nucleon collisions, and then try to fit the data by varying the parameters – the model will account for the data in an arbitrarily-chosen mathematical form (that is, C1).

[9] A reviewer pointed out, correctly, that the exclusion of causal considerations from phenomenological models or laws might not be completely accepted by everyone who participated in the debate. In particular, Cartwright (1983) connects phenomenological laws with what happens in concrete situations, and theoretical laws as describing no particular circumstance.

The second characteristic (C2) pertains to *how PMs are built*. Cartwright et al. (1995) claim that paradigmatic examples of phenomenological model-building are characterized by "independence from theory, in methods and aims" (p 148). In particular, PM-building is mostly based on phenomenological considerations[10], and on *ad hoc* varying of the mathematical conditions to fit the data adequately, where these 'moves' are not licensed by theory, nor follow from theory de-idealisation.

The third characteristic identifies the *origin of PMs* (C3), and was explicitly pointed out by Wilholt (2005), when defining PMs as models that are built starting from measurements and observations, with little theoretical input or information. He looks at the provenance of models as one reason to retain McMullin's distinction between theoretical and PMs, where PMs' history starts with a mathematical description of observed properties/behaviour.

Finally, a fourth characteristic (C4) is about the *goals of PMs*. Bokulich (2011) has more recently characterized PMs (built through *ad hoc* fitting to empirical data) as being useful for predictions, but not for explanations (C4).

MLMs, we claim, fulfil all conditions C1-C4: they are models derived from measurements through *ad hoc* adaptation of some mathematical construct for a given (non-explanatory) purpose. It is however important to emphasize that we are *not* claiming that MLMs *are* PMs. This claim has been already evaluated in the literature. For instance, Boge (2022) discusses a possible association between PMs and MLMs[11] (in particular, Boge argues for this in relation to deep learning models). He rejects the idea that these models are PMs, on the basis of disagreements on what PMs are, and of the challenges of associating a MLM to a given phenomenon. We have shown that, when it comes to C1-4, there is indeed agreement. However, addressing the second point will go beyond the scope of this article. For this reason, we just point out similarities between PMs and MLMs, and we use these as a basis for making specific claims on MLMs[12].

---

[10] These are 'phenomenological' in the sense of being purely descriptive lacking any theoretical justification (Wilholt 2005)

[11] As a reviewer pointed out, Pietsch (2022) draws a comparison between ML and 'phenomenological science' that is in agreement with what we do, even though he connects this 'phenomenological' science to causality, which is something that we avoid.

[12] It is interesting to point out that even Boge, while denying that deep learning models are PMs, still makes claims about the nature of such models on the basis of something that they have in common with PMs, in particular the idea of 'instrumental-qua-devoid of content' derived from Bokulich's work (2011)

Consider C3 first. A key step in building a MLM is to collect data sets for training. In science, these data sets are usually constructed from measurements[13] (except for certain types of synthetic data); therefore, the starting point for constructing MLMs - their *provenance* - is the same as the one of PMs: measurements.

C1 is also central in the construction of MLMs. As we have said, MLMs are mathematical representations of the statistical regularities that a ML system learns from its training data, and that it uses to formulate predictions on the target-phenomenon of interest. In other words, the goal of building a MLM is to describe the relation between selected features in a mathematical (and computational) form that allows, in the case of ML, predictive tasks like regression or classification. This is similar to the dynamics that McMullin describes for C1.

C2 is relevant too. The construction of MLMs is shaped by technical and engineering-related considerations that are justified by the proximate goal of 'selecting' the model that fits the data better. For instance, whether one chooses between discriminative (e.g. nonlinear kernels, decision tree, convolutional neural network, etc) or generative (e.g. variational autoencoders, transformers, etc) algorithms will not only depend on the problem, but also on the type of data one is dealing with. We mean this not in terms of the data domain (e.g. biomedical, financial, etc), but rather in the more technical sense of *data modality* (i.e., image, text, etc). Each ML algorithm will come 'pre-packaged' with a number of assumptions about the nature of the dataset it can be applied to, which are not related in any obvious way to the specificities of the context of implementation. Put it differently, MLMs describe a function mapping input labels to output labels. However, the mapping, *per se*, receives inputs especially from the mathematical nature of the used algorithmic tools (i.e., they are 'ad hoc'). For instance, in early cases of cancer genomics using support vector machines (SVM), classifiers were often built to distinguish between cancer-causing vs cancer-neutral somatic mutations. Those classifiers (see, e.g., Capriotti and Altman 2011) had continuous outputs from 0 to 1, where 0 was 'cancer-neutral' and 1 was 'cancer-causing'. Given the continuous values, thresholds for classification had to be chosen. But the choice of thresholds (e.g., 0.5) was usually motivated on the basis of technical considerations, and from the point of view of its 'theoretical' justification can be considered *ad hoc*.

[13] It is important to stress that we are talking about ML in science - in other contexts, the origin of data sets might not be 'measurements' (e.g., images of cats and dogs). One might also make the claim that certain medical imaging outputs are not strictly speaking measurements, and they are more akin to actual photographs (e.g., the pictures taken by a colonoscopy camera). These might be instances where the 'measurement lens' is a stretch, but in many cases imaging technologies require extensive measurement procedures.

Finally, central to MLMs are *predictions*. These are essential to measure the performance of models on the test set, and they are taken to be one of the things that ML can do well. Moreover, it is exactly the attention to 'predictive modelling' that has been seen as central for methods like ML, and the way these differ from more 'traditional' statistical modelling practices, which are focused explicitly on causality and explanation (Breiman 2001; Shmueli 2010). If we take Breiman's brief historical reconstruction to be roughly correct, a scientific community developing decision trees, neural nets, and SVM systems (that is, ML systems) started to grow from the mid-1980s in the context of speech recognition, image recognition, and handwriting recognition. According to Breiman, "[t]heir goal was predictive accuracy" (2001, p 205), and what mattered to them was algorithms' "strength as predictors, (...) and what gives them good predictive accuracy" (p 205). The centrality of predictions has been also stressed after Breiman. In addition to the often discussed article by Shmueli (2010), Bzdok et al (2018) have more recently contrasted 'traditional' statistics and ML, and claim that "ML concentrates on prediction by using general-purpose learning algorithms to find patterns in often rich and unwieldy data" (p 233). Daoud and Dubhashi (2023), while pointing out the existence of a third statistical hybrid culture, also take as a given that 'prediction' and 'predictive performance' are central in the *modus operandi* of contemporary ML. Rudin (2025) accepts the distinction between Breiman's two cultures - and implicitly the importance of prediction -, even though she disagrees on some implications of Breiman's analysis. While a few articles are certainly not enough to generalize, they illustrate in which sense predictions are central for the ML community, and how the alignment with C4 can hardly be overestimated. But one can argue against this in two ways[14]. First, one can say that there are fields (such as cognitive neuroscience) where MLMs, and deep learning in particular, can be used as scientific models to study, e.g., the brain connectivity (Cichy and Kaiser 2019). While this is certainly true, it can also be regarded as a non-prototypical use, which does not align with the typical usages of MLMs in the majority of other natural science disciplines, where the focus is instead strongly on prediction and predictive goals. But there is a second point, namely that now MLMs are used for a variety of scientific goals, beyond prediction (and this is not limited to cognitive neuroscience). For instance, in genomics (Watson 2022) the predictions made by MLMs are complemented by interpretable models, or explainable AI models, to unpack "the reasoning that underlies high-performance statistical models (...) [and] mine for insights and suggest novel hypotheses" (Watson 2022, p 1503). In this case, MLMs are contributing to other goals beyond prediction. Therefore, one can argue, predictions lose their centrality. But predictions still come first. In this regard, in particular, we can distinguish between a *proximate goal* and an *ultimate goal* of MLMs. A proximate goal is the immediate MLM's output, namely a prediction (or a classification, which is a kind of prediction). No matter what the MLM is used for, its immediate output (hence, its 'proximate goal') will always be a prediction, and this is valid for any ML system, from SVMs to LLMs (where the output is, strictly speaking, next-token prediction). The immediate outputs (i.e., the predictions) can then be used to achieve a goal (that we call 'ultimate goal'), which is usually set by the specific scientific context in which the MLM is implemented. Ultimate goals can vary significantly, and in the case of genomics Watson suggests that they are mostly about mechanistic insights or generation of new hypotheses. The ultimate goal is typically not achieved by MLMs alone, as the case of Watson shows: MLMs are one tool among many other tools. Interestingly, sometimes proximate and ultimate goals can coincide. For instance, the ultimate goal can simply be to classify data points in a data set into one or more categories of, say, biological entities such as genes, transcription start sites, etc (Libbrecht and Nobel 2015). In this case, the immediate output/proximate goal is the same as the ultimate goal. But, we realize, this is not the general rule. However, this does not change the fact that the proximate goal is always a prediction - and this fits nicely with C4.

[14] We thank a reviewer for pressing on these two issues

To sum up, the PM-lens can indeed shed light on the specificities of MLMs. First, given that algorithms are trained on data, the provenance of MLM is from measurements (at least in science). Second, ML models describe a dependency relation between labels by means of a carefully chosen mathematical form, as much as PMs do, and, as PMs, are built by resorting to *ad hoc* varying of mathematical conditions. Finally, in MLMs prediction is central, which is compatible with the predictive goal that is usually associated to PMs.

## 4. MACHINE LEARNING MODELS AND THE ROLE OF THEORY

Seeing MLM through the lens of PMs, we argue, is useful for understanding the theory-ladenness of MLM-building practices. Let us elaborate this in more detail.

Morrison (1999) distinguishes between aspects of *PM construction* from aspects of *PM use*. This distinction, in particular, is advocated to disentangle the complex and intricate relation between theory and PMs. Unlike in C1-C4, where there is agreement on the main points, this relation has been a topic of heated disagreements. For instance, McMullin (1968) takes an extreme position, by conceiving PMs as derived completely from measurement and being theory-free, while drawing a sharp separation from theoretical models, which are explanatory and theory-laden. Cartwright et al. (1995) see PM-building as *not theory-driven*, where this would grant a high-level of independence from theory. Wiholt (2005) claims that PMs are built with little theoretical input, even though the point is not developed in detail. But the relation between PMs and theory, Morrison says, is much more complicated than 'clear-cut' separations. PMs should not be seen as *completely* independent from theories: even though they provide a model of a phenomenon, and they seem to be based on fitting data to various mathematical formulations, they "can also be reliant on high level theory" (1999, p 46), especially in the way they are *applied*. For example, in discussing the model of the boundary layer describing the motion of a fluid, Morrison notices that two different theories are required to solve the hydrodynamic nonlinear equations: the fluid is divided conceptually into two regions, each requiring different approximations and different theoretical descriptions, such that the model "relies on two different theories for its applicability" (p 46).

In this section – and despite the differences between the context of the debate on PMs and the present context - we resume the suggestion of separating model construction from model use, and apply it to the analysis of the relation between domain-theory (understood in the sense described in Section 2) and MLM. We will not discuss 'model-use', since we tend to agree with the relevant literature: in order to use MLM in science for a number of different tasks (explorative, generative, etc), scientists typically rely on domain-theory. This is especially relevant when the 'ultimate goal' of a MLM is different from its proximate goal (see Section 3.2) - and given that the ultimate goal is conceptualized typically in domain-theory terms, then domain-theory plays a central role. But *model construction* rests especially on C1-4, and because of this, reveals ways in which MLMs and domain-theory relate (or not relate) that have been neglected in the literature, and that have crucial consequences for scientific practice and the methodology of AI-based science. Focusing on model construction and C1-4 will allow us to identify new categories of 'theory-ladenness' for MLM, which we define as follows:

- *Theory indifference*: a specific activity *x* in the process of *building* a model *m* within a domain *D* is *theory-indifferent* when no reference to domain-theory of *D* is necessarily required for the execution of tasks prescribed by *x*.
- *Theory-infection*: a model *m* is *theory-infected* with domain-theory *T*, when *T* is *implicitly* (i.e., outside the modellers' intentions and control) slipped into one of *m*'s components, but this is unrelated to *m*'s construction.

Before showing the dynamics underpinning theory-indifference and theory-infection in the practice of ML, two remarks about these definitions are in order.

First, as an attribute of MLM-building practice, theory-indifference is distinct from the three senses of theory-ladenness discussed in Section 2. If an activity is theory-indifferent, modelers are certainly not testing any theory. Moreover, they need not make any assumption that is informed by the theory, nor they need to bring any theoretical repertoire that will be used explicitly to execute or plan that activity (i.e., the activity is not theory-informed). Finally, domain-theory will not be used to formulate expectations of any sort (i.e., the activity is not theory-directed). However, it should be noted that theory can still inform, in some specific cases, but *it needs not to*: it is not explicitly and fundamentally required for implementing the process under consideration. The second remark is about the term *theory-infection*, which is here conceived as an attribute *of the model* rather than an attribute of the process of building or using a model. One might think that, if the process of building a model is theory-indifferent, then the model is theory-free. However, this is not necessarily (and also is not usually) the case; models can be 'infected' by theoretical priors in various ways, even if domain-theory plays no role in the model construction phase. But, as we will show, theory infection is still different from subtle senses of theory-ladenness like theory-informed.

In what follows, we illustrate more in detail the extent to which MLM construction is theory-indifferent (4.1, 4.3) and why, even if MLM can be theory-infected, the modeling practices remain theory-indifferent nonetheless (4.2).

### 4.1 The role of theory in the construction of machine learning models

To understand the level of theory-ladenness in the construction process of a MLM, and clarify the 'new' senses of theory-ladenness we previously introduced, it is useful to examine such a process in comparison with the construction of other kinds of models commonly used in scientific practice. Consider for example the well-known SIR model used in epidemiology to study the dynamic evolution of an infectious disease in a population (Milgroom 2023). The model relies on three *state variables*, namely $S$ (i.e. individuals in the population susceptible to contracting the infectious agent, such as a virus), $I$ (i.e. individuals in the population who have contracted the infectious agent and can transmit it), and $R$ (i.e. individuals in the population who have contracted the infectious agent and can no longer either contract or transmit it). In addition, the model introduces two parameters of precise biological significance, which are the *average infectious rate* $\beta$ (denoting how many susceptible individuals in the population get infected daily on average), and the *average removal rate* $\gamma$ (denoting how many infected individuals in the population recover or die daily on average). These variables and parameters are combined to obtain a compact mathematical description of the system's dynamic in terms of a system of three differential equations:

$$\frac{dS}{dt} = -\beta \frac{S}{N} I \qquad (1)$$

$$\frac{dI}{dt} = \beta \frac{S}{N} I - \gamma I \qquad (2)$$

$$\frac{dR}{dt} = \gamma I \qquad (3)$$

The specification of such an equation system is a theory-laden process where experimental data play a marginal role. Notably, this process can be qualified as both theory-informed and theory-directed. It is theory-informed because the choice of variables and functional dependencies to be considered are essentially the result of theoretical considerations based on domain-theory (i.e., in particular the framework) of the target phenomenon at stake, i.e. the spread of epidemics. The assumptions informing the equations are not induced from data but *they are the result of theoretical considerations* following from immunology. Moreover, the construction of the model itself is theory-directed, and based on a formulation of 'laws' or other 'models' coming from the 'theory' of immunology (understood in the sense of the third level of Douglas and Magnus' account) as well as the 'framework' (understood in the sense of the fourth level of Douglas and Magnus' account), and which create specific expectation on what is going to be observed. The only relevant task where data play a role is the estimation of the parameters $\beta$ and $\gamma$, which is based on experimental observations and measurements. However, note that this parameter estimation is usually obtained by performing experimental tasks designed on the basis of hypotheses that are drawn from theory (understood in the sense of the third level of Douglas and Magnus' account), hence it remains a theory-directed task in essence. The experimental estimation of parameters in 'traditional' statistical models (in opposition to the automatic learning of parameters in MLMs, as we will argue in the following sections) is a theory-informed task too. In this case, theory is indeed fundamentally required, because without theory one has no starting point for conceptualizing the relations between parameters and variables. Theory can thus be qualified as *necessary* because, in order to construct an accurate and reliable model of epidemics spread, the theory of the specific domains of epidemiology and immunology (in the sense of both the third and fourth level of Douglas and Magnus' account) is indispensable.

Different considerations emerge instead if we examine the construction process of MLMs, where essential aspects of this process are *theory-indifferent,* i.e., one need not to provide any interpretation of MLM components in terms of domain knowledge coming from the scientific context in which the MLM is intended to be used. Theory is *not necessary* because, to construct MLMs, theory (in the sense specified in Section 2) *can be* ignored. This is not to say that domain-theoretical considerations are always absent in scientific practice; rather, what we say is that they are *not necessary* to obtain accurate (in the technical sense of *in-distribution predictive accuracy* typically involved in ML practice) models, in contrast to what happens with more 'traditional' kinds of scientific models (e.g. the SIR model), which necessarily require domain-theoretical considerations for their specification. In the next section, we will show more in detail how the fundamental steps of MLM construction are substantially theory-indifferent.

*4.1.1 Theory-indifference in the selection of parameters and hyperparameters*

A MLM is specified by two sets of mathematical entities called *hyper-parameters* and *weight-parameters* (or simply *parameters*). Hyper-parameters are the parameters that determine the skeleton of the model, thereby constraining its possible final structure within certain given borders. The term is taken from Bayesian statistics, where a hyper-parameter is a parameter of the *prior distribution* fixing the set of possible *posterior distributions* that a model can fit (see, e.g., Bovens and Hartmann 2004). In the context of ML, the nature of hyper-parameters vary depending on the specific kind of architecture and framework considered. In the case of *deep neural networks* (Baldi 2021), for instance, hyper-parameters describe the topology of the network (fully connected, convolutional, recurrent, etc.), the kind of activation function used (linear, sigmoid, tan-h, etc.), etc. Beyond their specific nature, hyper-parameters play the same very specific role in all ML contexts, i.e., they fix the set of all possible models (i.e., predictive function/distribution) that the ML system can learn from the training data. Given a class of possible MLMs, determined by the hyper-parameters, the *actual model* is specifically determined by the *weight-parameters*.

The construction that leads to a MLM requires a specification of both the weight-parameters and the hyper-parameters. As we have seen in Section 3, the specification of weight-parameters is a fully-automated process that ultimately consists of solving an *optimisation task*, i.e., finding the minimum/maximum of a function accounting for some statistical magnitude (e.g., prediction error, variance, etc.) relative to the interaction between the model and the training sample – this is, in essence, the characteristic C1 that MLMs share with PMs. The performance of this optimization task requires no reference to the theoretical background of the specific domain to which the model is applied: it is a pure mathematical operation that, as such, can be qualified as *theory-indifferent*. This claim is supported by the fact, for instance, that the same optimisation functions and procedures can be exported and applied in different domains without requiring any theoretical adaptation. For example, the loss function *mean absolute error* can be identically applied in all the application domains that require learning a regression model, independently of whether this model describes the relation between *age* and *cancer risks* or the relation between the *financial hazard* and the *long-term income* of an economic agent. Theory-indifference is also evident if we consider the heuristic strategies used to implement the optimisation procedures that underlines the learning of weight-parameters. In non ML-based science, the heuristics that guide the scientific model-building process make a fundamental use of hypotheses that are formulated with the support of the existing corpus of domain-specific background knowledge. Consider the paradigmatic case of *decomposition* and *localisation*, two important heuristic strategies that guide the construction process of mechanistic models (Bechtel and Richardson 2010). Both rely on a fundamental contribution of domain-theory[15] (i.e., in particular, the 'framework' as understood in Section 2) for the formulation of hypotheses regarding the specific component-parts of a mechanism and the functions they perform. On the contrary, the heuristic strategies used in weight-parameters learning just exploit fundamental mathematical properties of the optimisations task they are supposed to solve – and this is related to the characteristic C2 that MLMs share with PMs. Consider in this regard the *stochastic gradient descent* described in Section 3. This heuristic strategy exploits a fundamental mathematical property of differentiable functions, which guarantees that minima can be effectively approximated by following the value path of its gradient: no domain-theory is required for the application of this heuristic strategy: all one has to know is the specification of the loss function. In other terms, stochastic gradient descent is *ad hoc* in the sense specified in Section 3.

Similar considerations hold for the hyper-parameters' specification. True, this is not always a fully automated task, and may require a suitable combination of automation and hand-made work. In general, a hand-made pre-selection of the hyper-parameters of the model is performed before the training phase, while automatic optimisation procedures (analogous in nature to those used for the learning of parameters) are typically used in validation to automatically fine-tune the hyper-parameter values on the training data distribution and avoid

[15] The contribution can be specified as a weakly directed theoretical contribution (Franklin 2005; Waters 2007)

notorious issues, such as overfitting. Domain-theoretical considerations can (and sometimes do) come into play in the hand-made preselection of the hyper-parameters. However, they are *not strictly necessary* to this task, *because*, even without any theoretical prior from the domain-theory, hyperparameters' specification can be performed accurately (i.e., leading to models that have good predictive performances according to the available metrics). Necessary to the hyper-parameters' specification are instead considerations of mathematical and engineering nature, e.g., related to the specific predictive task at stake (e.g., regression *vs* classification, as required by C4) and the format of the data to be processed (e.g., tabular data, time series, etc). These are *necessary because*, without resorting to those considerations, hyper-parameters' specification cannot be done properly. Again, attributing C2 to MLMs is central here, given that the emphasis is on '*ad hoc*' moves *not licensed* by domain theory - and this means that these are not even theory-informed. For example, in the analysis of time-series with neural networks, modelers typically select the recurrent topology due to its ability to support the processing of sequential data (Goodfellow et al. 2016), independently on whether these data represent fluctuations of energy market or brain signal. Similarly, convolutional topology for neural networks is commonly adopted in the analysis of images for its ability to filter and aggregate information from different regions of the image in parallel, independently on whether the images represent cats and dogs or nevus and melanomas. In all these cases, no considerations from domain-theory have to necessarily inform, let alone direct. In support of the claim that hyper-parameters specification is substantially a theory-indifferent task, we can also mention the increasing diffusion of fully automated procedures for the pre-selection of the hyper-parameters based on the application of meta-learning algorithms (Vanschoren 2019): these are optimisation procedures substantially analogous (and hence theory-indifferent) to those adopted in the learning of weight-parameters.

Before going any further, please note that the claim that the specification of parameters and hyperparameters is a theory-indifferent task does not imply that this task is *always* performed by modelers without mediation from the theory of the scientific domain of implementation. The ML literature is replete with examples of MLMs whose hyper-parameters have been selected *also* based on domain-theoretic considerations, or whose training is executed via optimisation algorithms opportunistically constrained with domain-specific background knowledge (and therefore partially *theory-informed*). A classic example is *AlphaFold*[16], the deep learning model developed by Google DeepMind for addressing the protein-folding protein. The hyper-parameters specification of this model is replete with theoretical considerations. Among the many theory-laden aspects, the topological structure of the AlphaFold network has been blueprinted explicitly considering the fact that the protein-folding process consists of three consequent prediction steps (primary-to-secondary, secondary-to-tertiary, and tertiary-to-quaternary structure of the protein), hence following explicit domain-theoretical considerations. However, what we claim here is different: considerations that appeal to domain-theory (understood in the encompassing sense described in Section 2), although they can sometimes be used in the hyper-parameters specification or to constrain the weight-parameters learning, they are *not necessary* for these tasks, as instead they are in more 'traditional' model-building practices. Predictively accurate MLMs can be constructed - and this is the common practice - with no reference to domain-theory at all.

[16] For an updated overview of the model's architecture, see (Yang et al. 2023)

Now, one might be tempted to argue that domain-theory still results essential for two tasks that are fundamental in MLM-building practice, notably the sampling and preparation of training data, and the so-called process of *features engineering* (Duboue 2020). The next sections analyse these two processes in more detail.

### 4.2 Domain-Theory in Database Curation, and Theory-Infection

As noticed in Section 3, MLMs are inferred from training data sets, which are collections of measurements (i.e., C3 applies here). It is well-known that 'measurements' - especially in contemporary science - are never 'direct' or 'raw' observations devoid of theory. This applies even more to data sets used in ML, which are highly processed and idealized versions of scientific measurements. As a consequence, domain-theory, in both forms corresponding to the third and fourth level of Douglas and Magnus' account, is already present in the data sets acquired to construct the training sample before any data processing procedure is done by ML specialists. In this regard, Leonelli (2016) has documented the epistemic subtleties behind the construction of databases in biology, in particular for what concerns data curation practices. For instance, "terms used for data classification should be the ones used by biologists to describe their research interests - that is, terms referring to biological phenomena" (Leonelli 2016, p 116). This means that a theoretical, pre-conceived understanding of biological phenomena is inevitably embedded in the data-sets that are drawn from biological databases, and the same substantially holds for any other discipline: database curation is a theory-informed process because domain-theory provides constraints on how the data must be curated. This reflects a widespread idea, according to which there is no such a thing as 'raw' or 'pure' data: theoretical priors are always an essential component of data sets. If domain-theory is a fundamental constituent of data, *a fortiori* it will have a necessary influence on the computational procedures that make use of these data sets, i.e., the training of the parameters and the fine-tuning of the hyper-parameters. From this, one may conclude that the 'blanket view' of theory-ladenness is still valid: domain-theory is necessarily slipped in the models, and hence MLM construction must be theory-laden. However, the consequences of data curation practices are not as straightforward as it may seem.

First, it should be noted that data collection and curation practices cannot be considered an integral part of the MLM-building process. The datasets on which ML systems are trained are indeed commonly prepared separately by data specialists who rarely coincide with the ML scientists responsible for MLMs' specification and training. Consider for example the dataset Siim-pneumothorax provided by the NIH[17]. This contains chest x-ray images with binary annotations indicating the presence or the absence of Pneumothorax in a sample of patients. The annotation is done by 'domain' (as in 'domain-theory') experts, but their involvement ends here. The dataset is publicly available for training diverse ML systems with different architectures, spanning from standard CNNs to Concept-Based models and transformers. Whether a particular ML architecture can be trained on this dataset depends on engineering considerations rather than domain knowledge. For instance an LSTM cannot be trained on Siim-pneumothorax simply because LSTM's are not designed to process images. The crux of the matter is that these architectural constraints are domain-agnostic, and any ML engineer can identify them without expertise in the domain of pneumology. The pattern exemplified here pervades the field: datasets are constructed by domain-experts and shared with ML researchers to train and compare various architectures. The ML community routinely benchmarks new architectures against existing ones (the 'state-of-the-art') using standardised datasets, a practice commonly mandated by all the major ML conferences (e.g., NeurIPS, ICML, ICLR, AAAI, etc.). This widespread approach shows the separation between domain experts (needed for annotation) and 'architectural expertise' (needed for model development).

This is consistent with Leonelli's work on data-intensive science (2016), which points to the neglected scientific figure of the *data curator*. These data curators develop a specific expertise which is distinct from the expertise of those who will use the databases. Hence, although data collection and curation practices are fundamental for building a MLM, they substantially *precede* the actual model-building process and are not properly part of it. The essential steps of MLM construction, i.e., parameters and hyperparameters specification, remain instead theory-indifferent, as previously argued in detail.

[17] https://www.kaggle.com/datasets/abhishek/siim-png-images

Nonetheless, theoretical priors used explicitly by data curators remain an essential component of training data and, as such, they are 'implicitly embedded' in the final MLM trained on them. One could ask: is this not enough to qualify MLMs as theory-laden? But this question is unclear, because reflecting about theory-laddenness in general terms could be misleading: there exist in fact different typologies of *theory-laddenness*, as we have shown in Section 2. Instead, the question should be reformulated, more properly, as: which kind of theory-ladenness are we dealing with in the case of MLMs? One can interpret it as another case of 'theory-informed' (Waters 2007). However, in theory-informed contexts theoretical priors are explicitly embedded in the model structure by modelers, and this makes such priors clearly identifiable within the final model structure. With trained MLMs, on the contrary, domain-theoretical priors embedded in the training data are *passively* inherited by the model as a consequence of the model training. This is not the only relevant difference. Indeed, for theory-directed and theory-informed models theoretical priors inherited from data can be easily identified within the model structure, e.g., by looking at the semantic interpretation of the model's parameters that we fit on such data. Consider again the SIR model previously examined. This model embeds two parameters in the equations, which are assumed to represent (i.e., their semantic interpretation is) the specific rates of variations in the number of infected and removed subjects during a certain time-window of the epidemics spread. To fit these parameters, scientists refer to properly curated data samples, whose features are assumed to represent the evolution of the number of infected, respectively, removed subjects in the given temporal window of interest. Information in data is commonly structured according to a "stocks-and-flows" ontology, which is common in data-collections referring to the temporal evolution of dynamical systems (Sterman 2000). In particular, in the case of data for the SIR model, stocks usually represent the number of infected/healthy/removed subjects at each given time and flows represent the latter instant variations (i.e., their derivatives with respect to time). These ontological considerations constitute a strong theoretical prior, which one can clearly identify in the model's structure simply by considering the semantics of the model's parameters, which precisely denote the instant variations in time of the number of infected/healthy/removed subjects. Analogous considerations do not hold in the case of MLMs. In many cases, the parameters of certain MLMs (particularly in deep neural networks) lack in fact any semantic (mediated by domain-theory) interpretation understandable to human users, making the models *semantically opaque* in the terminology introduced by Facchini &

Termine (2021). Furthermore, these parameters do not denote any real-world component or phenomenon—that is, they are *not elementwise representational*, as Freisleben et al (2024) point out. The parameters of those MLMs possess, in a sense, a pure instrumental meaning: they are parameters that allow the trained model to minimise the predictive error over the training data. The information inherited from the data is scattered across millions — or even billions — of parameters that lack any comprehensible semantics, making it virtually impossible to reconstruct the theoretical contributions that the model has inherited from data.

This is an unusual situation: MLMs are theory-laden because priors are passively inherited, but domain-theory plays no clear role. This is why we coin the term *theory-infection*: MLM, are indeed theory-laden, but not in any of the senses discussed in the literature.

**4.3 Features engineering: from theory-directed to theory-indifferent**

The second step where domain-theory seems to play an important role is features engineering, i.e., the construction process of the features composing the training sample. This process is based on the collection of large amounts of observational data, which are finely pre-processed and sampled to obtain raw features. These data may come either from databases (as discussed in the previous paragraph), or from more direct measurements taken by a given scientific group. These are then subjected to various manipulation processes directly by ML specialists, which ultimately result in derived features that are used as input for training procedures. Domain-theory seems to play a non-negligible role in the process of feature construction. After all, deciding which variables of a target-phenomenon to consider for predictive purposes, and how to combine them in suitable representation formats, is a task that requires an extensive knowledge of the phenomenon under investigation. This is certainly true for more traditional - and older - kinds of MLMs, like decision-trees or random forests, which operate with hand-made features.

However, the advent of automatic feature learning algorithms, whose operation is essentially based on the execution of optimisation tasks similar to those used to train MLMs, are gradually eliminating any role for domain theory in the feature construction process. Examples of this path from theory-informed to theory-indifferent features engineering can be found in various domains of scientific investigations using deep learning systems. These systems are capable of generating predictions directly from 'raw-data' (e.g., *images*), and incorporate features engineering as a step of the predictive inferences they perform.

To illustrate and exemplify these considerations on features engineering, let us consider the case of MLMs in neuroimaging-based psychiatric research (Eitel et al. 2023). The detection of psychiatric disorders is a notoriously challenging task because the underlying mechanisms of these pathologies, with a few exceptions such as Alzheimer's disease, remain mostly unknown or only partially understood. This makes traditional 'theory-directed' or 'theory-informed' modelling techniques, such as mechanistic models and simulations, difficult to apply. On the contrary, MLMs have proven to be easier to implement, in particular due to the independence of their training from theoretical considerations.

The analysis of literature (see, e.g., Eitel et al. 2023) not only shows that neuroscientific theory has a limited influence on the MLM construction process applied in this domain, but it also displays a trend towards an increasing theory-indifference of all model-building steps, including features engineering. This is particularly evident, again, in the shift from more classical ML architectures (e.g., decision-trees), which require the use of hand-crafted high-level features, to deep learning systems, which can instead learn their features directly from raw-data (see, Fig. 3) in a fully automatic manner, akin to C2. Let us clarify this point a bit more in detail.

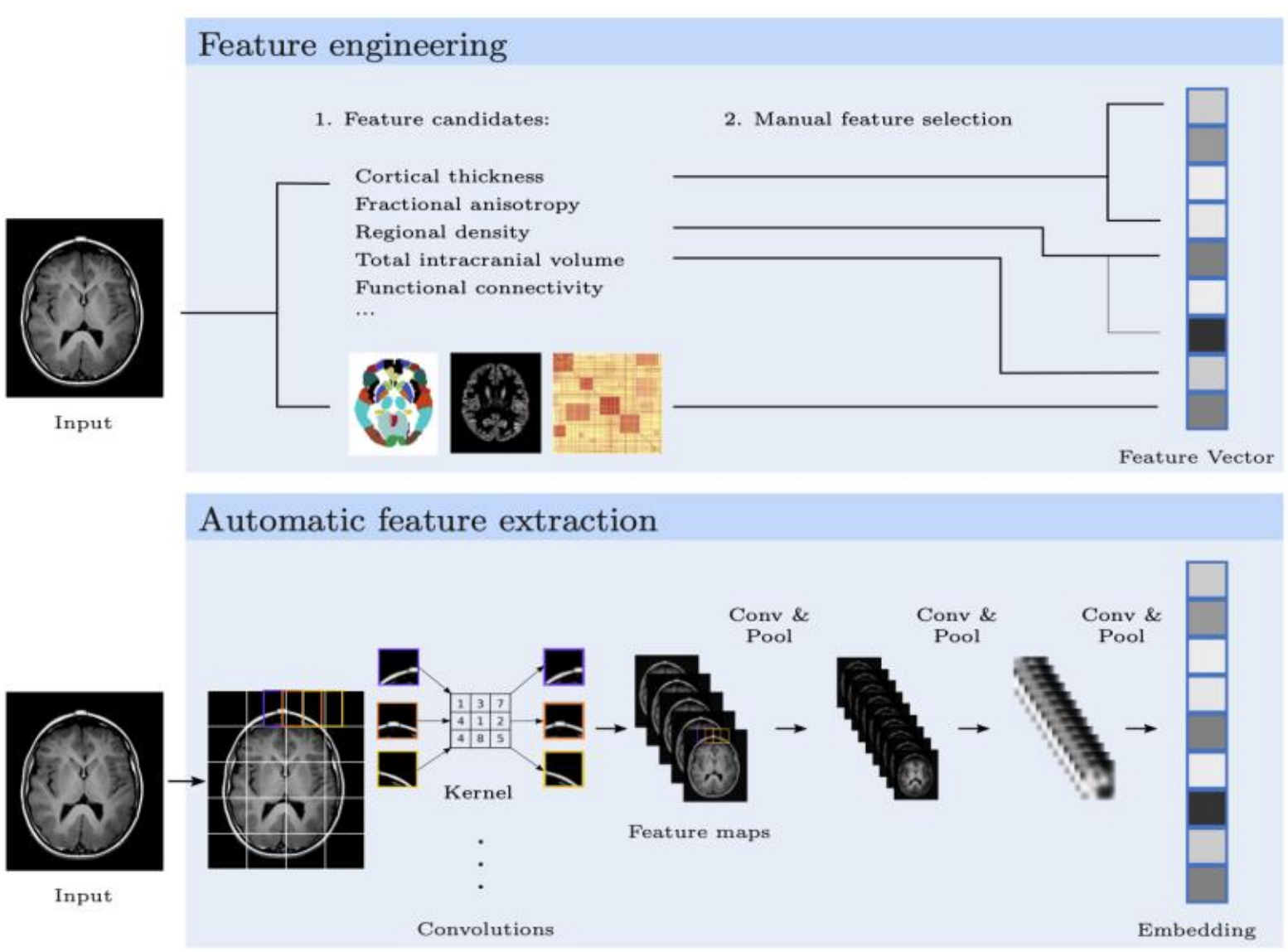

Figure 3 (from Eitel et al. 2023)

Both for 'classical' and deep learning models the model-building process starts with the collection of brain images, which are here represented as grids of pixels encoded in the form of numerical matrices. Each cell of the matrix (i.e., each pixel) represents a single low-level 'raw' feature. These features must be converted into high-level features that allow for predicting the target psychiatric disorder. This is, in substance, the *features engineering* process. Differences between 'classical' and more contemporary deep learning models emerge exactly at the level of this process. Typically, scientists select and extract manually from images a number of high-level variables (features), such as *cortical thickness, fractional anisotropy* etc., and thus use ‘raw-data’ to determine their values. The choice of the variables depends on explicit domain-theoretical considerations (it is weakly theory-directed): for example, modelers focus on *cortical thickness* because they are aware (from domain-theory) of the relevance of this feature for the prediction of specific psychiatric disorders. In more contemporary MLMs things go differently.

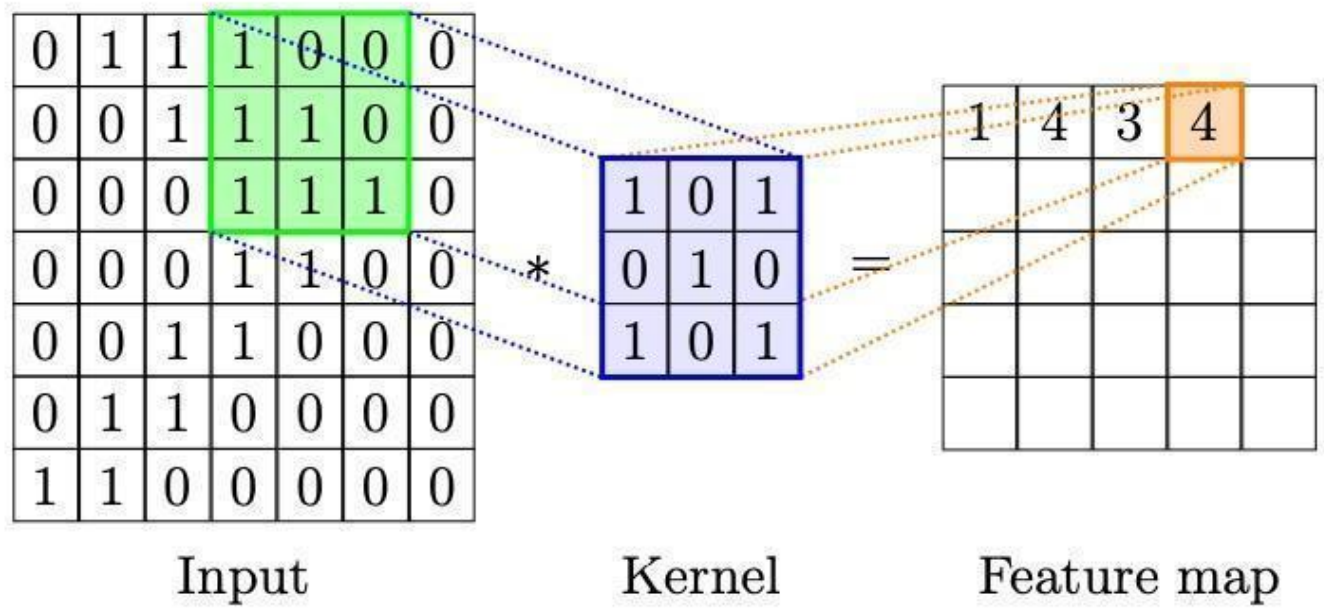

Figure 4: Representation of a 2-dimensional convolution operation (image from Eitel et al. 2023)

Consider for example the case of *Convolutional Neural Networks* (CNN). CNNs do not require hand-made high-level features but are able to automatically extract these features from ‘raw-data’ through the application of a mathematical operation known as *convolution* (Fig. 4). This is a linear transformation based on the application of a kernel of parameters to the input. The various regions of the input are processed through a filter unit that computes the weighted sum of the input-features (i.e., the pixels of the image encoded in the matrix) in the region and kernel parameters, hence mapping the result into a feature map. In general, data are processed through iterative convolution operations, which eventually produce the high-level features the model uses for predicting the target, technically called ‘embeddings’. Differently from hand-made high-level features, embeddings may not possess a clear interpretation and represent magnitudes of the input that do not possess any clear meaning for the domain-experts. Furthermore, and more importantly, the construction and selection of embeddings do not require any theoretical considerations related to the specific application domain. On the contrary, they rely only on the execution of pure mathematical operations based on numerical parameters, which are learned via standard optimization procedures analogous to those used for weight-parameters learning. With CNNs, feature engineering can be therefore qualified as a theory-indifferent process. This consideration can be generalized to the majority of the deep learning architectures used in the various domains of scientific research (Baldi 2021), including *foundation models* and related *transformers architectures* underlying contemporary Generative AI tools. In general, we can say that a trend exists in the ML community towards the increasing use of theory-indifferent automatic features extraction procedures, thereby eliminating any necessary dependence of the MLM-building process on domain-theory and contributing to make it a completely theory-indifferent activity.

## 5. CONSEQUENCES OF OUR ANALYSIS

What emerged from this analysis is that the role of domain knowledge in model construction in ML seems to be very limited: MLM construction looks indeed mostly (and increasingly) a theory-indifferent activity, where no reference to domain-theory is necessarily required in the various steps of the MLM-building process. But what consequences should we draw from this analysis? Here we discuss two far-reaching consequences.

The first consequence pertains to the differences between ML strategies and other modelling strategies. If the blanket-view of theory-ladenness applies to ML modelling practices as it applies to other modelling strategies, then those siding in favour of a continuity between ML strategies and other modelling strategies would take this as an additional reason to support continuity itself. However, in the case of ML, we have shown that, even if in a small number of cases theoretical considerations can play important roles in MLM-building (Hansen and Quinon 2023; Gross 2024; Andrews 2025), this is *not necessarily* the case: one can construct a MLM with optimal performances in terms of the standardly adopted metrics[18] without making any reference to the domain theory. The fact that domain-theory is not necessary marks, we argue, an important discontinuity between MLM-building and other modeling strategies used in scientific research. This can be appreciated by emphasizing the *as-is* transferability of ML architectures and MLMs construction practices across different domains as a direct consequence of theory-indifference. By 'as-is', we mean that a MLM can be exported from one domain to a different domain without either re-adapting the model's inner structure to the theoretical background of the new domain, or justifying the model's implementation in the new domain on the basis of relevant domain-theory. For exporting successfully a ML architecture the only thing we need is a new training sample on which to re-train the system (i.e., on which to automatically adjust its weight-parameters and fine-tune its hyper-parameters). To better understand this point, consider a convolutional deep neural network (call it *Netty*) for image recognition as the one depicted in Fig. 3. Suppose Netty is initially trained to predict potential neurological symptoms of Alzheimer's using a sample of neuroimages with a given resolution, and thus achieves a certain desirable accuracy on test. Now imagine that the ML scientist responsible for building Netty is asked to build another ML predictive model for detecting signs of arthritis in the knee, using MRI-produced images with a resolution and format similar to the neuroimages used to train Netty. Without the necessity to advance any theoretical consideration about the new domain of application, the ML scientist will take Netty and re-train it on the new training sample of knee images. Hence, if they find some issues in the predictive accuracy of the re-trained model on the new dataset, perhaps due to slight differences in the format or resolution of the new images, they will perform an adjustment of the hyperparameters. Again, this operation will be arguably done without any reference to domain theory but will just appeal to mathematical and/or engineering-related considerations (*per* C2). What makes this possible is the substantial 'agnosticism' of ML architectures and related model-building practices with respect to different domain theories. In other words, a convolutional neural network with a certain structure can work well for all images with a similar format, regardless of what they represent (and therefore regardless of the domain from which they come). This, we claim, is an almost *unique character* of MLM construction practices[19], which differentiate them from the other typologies of models usually involved in scientific practice, and it is a direct result of the thesis of theory-indifference.

[18] In particular, we refer here to the standard notion of *predictive accuracy* (i.e., rate of correct predictions) measured on *independently and identically distributed data*, i.e., data that share the same underlying distribution of the data in the training sample (see, Schölkopf et al. 2021).

One could certainly counter-argue to this claim by pointing out that exportability across different domains is common also with other kinds of scientific models. For instance, philosophers of science have been debating the transferability of scientific models across different domains (Herfeld 2024). However, in most cases of model transfer, there is a significant amount of work that needs to be done to adapt the model to the new context, and this requires the use of domain theory, especially in the form of the framework as specified in Section 2. Consider a recent adaptation of the SIR model to analyse risk contagion among financial players (Aliano et al. 2024). In this work, the authors show that a SIR model can effectively describe the dynamics of risk contagion and propagation among financial players, provided that the variables of the model are interpreted as representing individuals subject to-, infected by-, or immune-to financial risk. We can be tempted to claim that the SIR model is nothing but a powerful mathematical instrument that can be easily re-adapted to different contexts by providing the opportune semantic translation of the variables involved and the re-tuning of the model's parameters. However, things are not so simple. In order to export the SIR model from the field of epidemiology to that of financial risk analysis, researchers must assume that the two phenomena (epidemics and the contagion and propagation of financial risk) have analogous dynamics, i.e. that they behave very similarly over time, so that the theoretical considerations from the field of epistemology that were used to develop the original SIR model also apply to financial risk contagion. This represents a strongly theory-directed assumption that can be advanced only by an expert in financial risk analysis, with an extensive knowledge of the dynamics of financial risk contagion. Things are instead radically different for MLMs, whose translation from one domain to another do not require any domain-theoretical expertise but only considerations concerning the format and structure of the data to be analysed. As said before, a neural network for image classification work can be equivalently applied to either distinguish images of *cats* and *dogs*, or of *naevus* and *melanoma*. The only operation required to export the model from one domain to the other is the retraining of weight-parameters, which is a completely theory-indifferent activity, as we extensively argued in the previous sections. On the other hand, the only kind of considerations required to perform this model-exportation concerns the format and accuracy of the data involved: a neural network constructed for classifying images cannot clearly be applied to tabular data, as well as its performances can change if the granularity and accuracy of the data involved is different.

[19] The only other case of as-is transferability seems to be network science, as explained by Humphreys (2019). In these cases, formal network models are indeed used to model a number of different domains.

One could also say that it is not new either that practitioners move from one domain to another. For instance, the history of molecular biology or bioinformatics is characterized by physicists migrating to biological research projects (Kay 2000; Stevens 2013). But what is happening in ML-based science is different. As in the case of model transfers, in all cases of practitioners migrating, there is a significant amount of *theoretical* work (viz, pertaining to domain knowledge) to adapt to the new field. A classic example is Gamow's contribution to biology. As a physicist, he pursued the biological question of the relation between DNA and amino acids by using the tools of cryptography (Kay 2000): famously, he hypothesized that the problem of the relation between DNA-amino acids could be treated as a coding problem. But he could not just use his expertise in cryptography *as-is*: in fact, his ideas and expertise had to be painfully adapted to the specificities of the biological domain. But if the thesis of theory-indifference is true, then transferability and migration in ML works differently than in the cases mentioned above. Given that ML practitioners can build models without knowing anything about the domain of implementation, then their expertise and practice, *seamlessly* can potentially travel from one domain to another *as-is*. If all of this is true, then one question that can be asked is how ML practitioners should be trained: do we need ML curricula specific for a given discipline, or just one, general-purpose curriculum, that will be adequate for future ML practitioners to work in any scientific discipline?

The second consequence of our analysis pertains to the debate on theory-ladenness itself, independently of the previous point. Our results suggest that there is an underexplored direction of the debate on theory-ladenness in ML (or, similarly, Big Data and data science). In particular, our results can shift the debate from a mere descriptive discussion, to a *normative* one. We have shown that theory-ladenness is not necessary in MLM construction, and that ML practitioners have a choice in deciding whether to infuse ML systems with domain-theory. This raises a number of questions: is it desirable to infuse domain-theory in ML systems? Are ML systems explicitly designed to account for domain-theory considerations better than systems that are not? This goes back to Cox's criticism of the algorithmic modelling culture when, in commenting Breiman's article, argues that the output of ML systems "without some understanding of underlying process and linking with other sources of information [i.e., domain knowledge], becomes more and more tentative" (Cox 2001, pp 216-217). A new (and old at the same time) question then would be: is this true? AlphaFold has already become a classic case study in philosophy of science, but it is typically used just to show the subtleties of how a ML system can be theory-laden, and not whether its theory-ladenness is desirable.

We do not have the space to provide an argument in favour or against theory-ladenness from this normative perspective, and we plan to do this in future work. However, for the time being we draw the attention to the fact that the ML research community is already experiencing a conflict between two different 'cultures' of ML modelling. On the one hand, an increasing effort is dedicated to the construction of general-purpose MLMs that can be easily adapted to a variety of scientific domains and tasks via a simple fine-tuning of their parameters and hyper-parameters, or even just by prompting them properly. Contemporary research on fully-automated AI is an extreme and clear example of this tendency (see, e.g., Lu et al. 2024). What supporters of this paradigm aim for is the construction of large-scale ML architectures (sometimes called *foundation models*) that automate scientific discovery and modelling, from the initial exploratory research hypotheses to the draft of the final paper. This implies the creation of domain- and task-agnostic models, which can be applied to do research on virtually any domain of scientific investigation. What a scientist has to do is just to prompt the ML architecture properly and fill it with the relevant data; the system thus fine-tunes its inner function and adapts to the specific domain and task of interest in a fully-automatic way. On the other hand, there is an increasing group of scholars and scientists arguing that the theory-indifference of MLMs-construction practice is leading to detrimental consequences, and the time has come for more verticalization and for a step back to domain-anchored models. One major argument advanced in this regard is that embedding domain-theoretical priors in the hyper-parameters specification and constraining the parameters learning is extremely useful, for instance, to make models more *robust* and increase their generalisability in so-called *out-of-distribution* samples, i.e., samples whose data have a distribution that differs significatively from the distribution of the training sample (see, e.g., Pearl 2019, Schölkopf et al. 2021, Kaddour et al. 2022). But this conflict on the role of scientific domain knowledge in constructing MLMs (as noted above) is not entirely new. In a sense, indeed, we can regard it as a continuation of the decennial conflict between the two cultures of statistics analysed in Breiman 2001's influential paper.

## 6. CONCLUSION

In this article, we have proposed an in-depth analysis of the relation between MLMs and the domain-theory of the scientific context in which they are implemented. Looking at MLMs through the lens of the debate on PMs, we have identified new dimensions of theory-ladenness. We have confirmed what most of the literature says on the theory-ladenness of how MLMs are used, but we have argued against a blanket view of theory-ladenness that also covers the construction of MLM, which is instead a substantially theory-indifferent process (especially for contemporary deep learning models), in the sense that theory is not necessarily required in any proper step of the construction process of MLM models. Based on these considerations, we have discussed two far-reaching consequences for the thesis of theory-indifference, and suggested a new direction for the debate on theory-ladenness of ML. Specifically, what needs to be argued for is whether an explicit reference to domain-theory in MLM-construction should be required to address other epistemic desiderata of MLMs different from usual predictive accuracy, such as explainability, robustness and generalisability. We do not have the space to discuss this normative issue, but we see it as being at the top of the list of priorities of debates on the epistemic significance of ML in the sciences.

**Declarations**

- **Availability of data and material:** not applicable
- **Competing interests:** not applicable
- **Funding:** Innovate UK Award Number 10063119 (ER), B4EAI SUPSI Internal Grant (AT, AF), SNFS Agorà Grant LAAGP_223930/1 "AI Meets Education" (AT)
- **Author's contributions:** names in order of contribution

- **Acknowledgement:** We are grateful to Max Jones, Ross Pain, and Ana-Maria Cretu from the University of Bristol for their comments on a very early draft of this manuscript. We thank Lena Kastner's group, the META Group at Politecnico di Milano, and Eran Tal for valuable feedback, as well as audiences at the conference 'Model, Representation, and Computation' in Paris and the PSA 2024 for important suggestions. We are indebted to two anonymous reviewers for feedback that substantially improved this article.